\documentclass[Afour,sagev,times]{sagej}

\usepackage{moreverb,url}
\usepackage{algorithm}
\usepackage{algorithmic}
\usepackage{color}
\usepackage[colorlinks,bookmarksopen,bookmarksnumbered,citecolor=red,urlcolor=red]{hyperref}
\newcommand{\etal}{\textit{et al.~}}
\newcommand\BibTeX{{\rmfamily B\kern-.05em \textsc{i\kern-.025em b}\kern-.08em
T\kern-.1667em\lower.7ex\hbox{E}\kern-.125emX}}

\def\volumeyear{2016}

\begin{document}

\runninghead{Cai and Yin}

\title{A sparsity augmented probabilistic collaborative representation based classification method}

\author{Xiao-Yun Cai\affilnum{1} and He-Feng Yin\affilnum{2}}

\affiliation{\affilnum{1}Zhenjiang College, Zhenjiang, China\\
\affilnum{2}School of Internet of Things Engineering, Jiangnan University, Wuxi, China}

\corrauth{Xiao-Yun Cai, Zhenjiang College, 
No. 518 Changxiangxi Avenue, College Park, 
Zhenjiang 212028, China.}

\email{dysfcxy@126.com, 7141905017@vip.jiangnan.edu.cn}

\begin{abstract}
In order to enhance the performance of image recognition, a sparsity augmented probabilistic collaborative representation based classification (SA-ProCRC) method is presented. The proposed method obtains the dense coefficient through ProCRC, then augments the dense coefficient with a sparse one, and the sparse coefficient is attained by the orthogonal matching pursuit (OMP) algorithm. In contrast to conventional methods which require explicit computation of the reconstruction residuals for each class, the proposed method employs the augmented coefficient and the label matrix of the training samples to classify the test sample. Experimental results indicate that the proposed method can achieve promising results for face and scene images. The source code of our proposed SA-ProCRC is accessible at \url{https://github.com/yinhefeng/SAProCRC}.
\end{abstract}


\maketitle

\section{Introduction}
Image recognition remains one of the hottest topics in the communities of computer vision and pattern recognition. During the past decade, sparse representation has been successfully applied in various domains. In face recognition, the pioneering work is the sparse representation based classification (SRC) \cite{R1}. Concretely, SRC employs all the training samples as a dictionary, and a test sample is sparsely coded over the dictionary, then the classification is performed by checking which class yields the least reconstruction error. SRC can achieve promising recognition results even when the test samples are occluded or corrupted. To further promote the robustness of SRC, Wang \etal \cite{R2} proposed a correntropy matching pursuit (CMP) method  for robust sparse representation based recognition. CMP can adaptively assign small weights on severely corrupted entries of data and large weights on clean ones, thus reducing the effect of large noise. Wu \etal \cite{R3} presented a gradient direction-based hierarchical adaptive sparse and low-rank (GD-HASLR) algorithm to tackle the real-world occluded face recognition problem. Gao \etal \cite{R4} developed a robust and discriminative low-rank representation (RDLRR) method by exploiting the low-rankness of both the data representation and each occlusion-induced error image simultaneously. Keinert \etal \cite{R5} designed a group sparse representation-based method for face recognition (GSR-FR) which introduces a non-convex sparsity-inducing penalty and a robust non-convex loss function.

Apart from classifier design, feature extraction is also a crucial stage in image recognition. The most classic subspace learning based approaches are principal component analysis (PCA) and linear discriminant analysis (LDA). Motivated by the recent development of sparse representation, Qiao \etal \cite{R6} presented a dimensionality reduction technique called sparsity preserving projections (SPP). To make SRC efficiently deal with high-dimensional data, Cui \etal \cite{R7} proposed an integrated optimisation algorithm to implement feature extraction, dictionary learning and classification simultaneously. To tackle the corrupted data, Xie \etal \cite{R8} explored a dimensionality reduction method termed low-rank sparse preserving projections (LSPP) by combining the manifold learning and low-rank sparse representation.

Recently, sparse representation has been applied to a wide range of tasks. Zhang \etal \cite{R9} developed a structural sparse representation model for visual tracking. Liu \etal \cite{R10} introduced the convolutional sparse representation (CSR) into image fusion. Guo \etal \cite{R11} proposed a sparse and dense hybrid representation-based target detector (SDRD) for hyperspectral imagery (HSI).

Another critical issue in sparse representation is how to solve the $\ell_1$-norm constraint problem. Zhang \etal \cite{R12} presented a survey of sparse representation algorithms and found that Homotopy and ALM can achieve better recognition performance and have relatively lower computational cost.

Akhtar \etal \cite{R13} revealed that sparseness explicitly contributes to improved classification. And they proposed a sparsity augmented collaborative representation based classification (SA-CRC) which employs both dense and sparse collaborative representations to recognize a test sample. However, CRC \cite{R14} utilizes all the training samples to represent the input test sample, which neglects the relationship between the test sample and each of the multiple classes. To overcome the drawback of SA-CRC, first we obtain a dense representation by probabilistic collaborative representation based classification (ProCRC) \cite{R15}, then we augment the representation of ProCRC with a sparse representation to further promote the sparsity of ProCRC. Moreover, different from conventional representation based classification methods that use class-wise reconstruction error for classification, we utilize the label matrix of training data and the augmented coefficient of a test sample for final classification. The proposed method is termed as sparsity augmented probabilistic collaborative representation based classification (SA-ProCRC). {\color{red}In summary, our contributions are as follows,
\begin{itemize}
\item We promote the sparsity of ProCRC by augmenting the representation of ProCRC with a sparse representation.
\item We employ an efficient classification rule to recognize the test sample, in which the explicit computation of residuals class by class is avoided.
\item Experimental results on diverse datasets validate the efficacy of our proposed method.
\end{itemize}}

\section{Related work}
Given $n$ training samples belonging to $C$ classes, and the training data matrix is denoted by $\mathbf{X}=\left[\mathbf{X}_{1}, \mathbf{X}_{2}, \ldots, \mathbf{X}_{C}\right]=[\boldsymbol{x}_1,\boldsymbol{x}_2,\ldots,\boldsymbol{x}_n] \in \mathbb{R}^{m \times n}$, where $\mathbf{X}_i$ is the data matrix of the $i$-th class. The $i$-th class has $n_i$ training samples and $\sum_{i=1}^Cn_i=n$, $i=1,2,\ldots,C$, $m$ is the dimensionality of vectorized samples.
\subsection{Sparse representation based classification}
In SRC \cite{R1}, a test sample $\boldsymbol{y}\in\mathbb{R}^m$ is firstly represented as a sparse linear combination of all the training data, then the classification is performed by checking which class leads to the least reconstruction error, the objective function of SRC is formulated as,
\begin{equation} 
\label{eq:obj_src}
\underset{\boldsymbol{\alpha}}{\textrm{min}} \ \left \| \boldsymbol{\alpha} \right \|_1, \ \textrm{s.t.} \ \left \| \boldsymbol{y}-\mathbf{X}\boldsymbol{\alpha} \right \|_2^2\leq \varepsilon
\end{equation}
where $\varepsilon$ is a given error tolerance.
When we obtain the coefficient vector $\boldsymbol{\alpha}$ of $\boldsymbol{y}$, the test sample $\boldsymbol{y}$ is classified according to the following formulation,
\begin{equation}
\label{eq:rule_src}
\textrm{identity}\left( \boldsymbol{y} \right)=\arg \underset{i}{\mathop{\min }}\,\left\| \boldsymbol{y}-{{\mathbf{X}}_{i}}{{\boldsymbol{\alpha} }_{i}} \right\|_{{\color{red}2}}
\end{equation}
where ${\boldsymbol{\alpha} }_{i}$ is the coefficient vector that corresponds to the $i$-th class.
\subsection{Collaborative representation based classification}
SRC and its extensions have achieved encouraging results in a variety of pattern classification tasks. However, Zhang \etal \cite{R14} argued that it is the collaborative representation mechanism rather than the $\ell_1$-norm sparsity that makes SRC powerful for classification. And they presented collaborative representation based classification (CRC) algorithm, which replaces the $\ell_1$-norm in SRC with the $\ell_2$-norm constraint, the objective function of CRC is formulated as follows,
\begin{equation}
\label{eq:obj_crc}
\underset{\boldsymbol{\alpha} }{\mathop{\min }}\,\left\| \boldsymbol{y}-\mathbf{X}\boldsymbol{\alpha}  \right\|_{2}^{2}+\lambda {{\left\| \boldsymbol{\alpha}  \right\|}_{2}^2} 
\end{equation}
CRC has the following closed-form solution,
\begin{equation}
\label{eq:solu_crc}
\boldsymbol{\alpha}=(\mathbf{X}^T\mathbf{X}+\lambda \mathbf{I})^{-1}\mathbf{X}^T\boldsymbol{y}
\end{equation}
where $\mathbf{I}$ is the identity matrix. Let $\mathbf{P}=(\mathbf{X}^T\mathbf{X}+\lambda \mathbf{I})^{-1}\mathbf{X}^T$, one can see that $\mathbf{P}$ is determined by the training data matrix $\mathbf{X}$. Therefore, when given all the training data, $\mathbf{P}$ can be pre-computed, which makes CRC very efficient. CRC employs the following regularized residual for classification,
\begin{equation}
\label{rule_crc}
\textrm{identity}\left( \boldsymbol{y} \right)=\arg \underset{i}{\mathop{\min }}\,\frac{\left\| \boldsymbol{y}-{{\mathbf{X}}_{i}}{{\boldsymbol{\alpha} }_{i}} \right\|_{2}}{\left \| \boldsymbol{\alpha}_i \right \|_2} 
\end{equation}
\subsection{Probabilistic CRC}
Inspired by the work of probabilistic subspace approaches, Cai \etal \cite{R15} explored the classification mechanism of CRC from a probabilistic perspective and developed a probabilistic collaborative representation based classifier (ProCRC), and the objective function of ProCRC is formulated as,
\begin{equation}
\label{eq:obj_procrc}
\underset{\check{\boldsymbol{\alpha}} }{\mathop{\min }}\,\left\| \boldsymbol{y}-\mathbf{X}\check{\boldsymbol{\alpha}}  \right\|_{2}^{2}+\lambda {{\left\| \check{\boldsymbol{\alpha}}  \right\|}_{2}^2}+\frac{\gamma}{C}\sum_{i=1}^{C}\left \|  \mathbf{X}\check{\boldsymbol{\alpha}}-\mathbf{X}_i\check{\boldsymbol{\alpha}}_i\right \|_2^2 
\end{equation}
where $\lambda$ and $\gamma$ are two balancing parameters. One can see that ProCRC is reduced to CRC when $\gamma=0$.
Suppose $\mathbf{X}_i^{'} $ is a matrix that has the same size as $\mathbf{X}$, and $\mathbf{X}_i^{'} $ only contains the samples from the $i$-th class, namely $\mathbf{X}_i^{'}=\left [ \boldsymbol{0},\ldots,\mathbf{X}_i,\ldots,\boldsymbol{0} \right ] $.  Let $\bar{\mathbf{X}}_i^{'}=\mathbf{X}-\mathbf{X}_i^{'} $, after some deductions, we can obtain the following closed-form solution to ProCRC,
\begin{equation}
\label{eq:solu_procrc}
\check{\boldsymbol{\alpha}}=\mathbf{T}\boldsymbol{y}
\end{equation}
where $\mathbf{T}=(\mathbf{X}^T\mathbf{X}+\frac{\gamma}{C}\sum_{i=1}^{C}(\bar{\mathbf{X}}_i^{'})^T\bar{\mathbf{X}}_i^{'}+\lambda \mathbf{I})^{-1}\mathbf{X}^T $ and $\mathbf{I}$ is the identity matrix.
\section{Sparsity augmented ProCRC}
In our proposed SA-ProCRC, the dense representation of ProCRC is augmented by a sparse representation computed by OMP \cite{R16}, and the optimization problem for sparse representation is given by,
\begin{equation} 
\label{eq:obj_omp}
\min _{\boldsymbol{\hat{\alpha}}}\left\|\boldsymbol{y}-\mathbf{X} \boldsymbol{\hat{\alpha}}\right\|_{2}, \text { s.t. }\left\|\boldsymbol{\hat{\alpha}}\right\|_{0} \leq k
\end{equation}
where $k$ is the sparsity level.

The augmented coefficient $\stackrel{\circ}{\boldsymbol{\alpha}}$ can be obtained according to the following formulation,
\begin{equation} 
\label{equ:coeff_fused}
\stackrel{\circ}{\boldsymbol{\alpha}}=\frac{\hat{\boldsymbol{\alpha}}+\check{\boldsymbol{\alpha}}}{\left\|\hat{\boldsymbol{\alpha}}+\check{\boldsymbol{\alpha}}\right\|_{2}}
\end{equation}
where $\hat{\boldsymbol{\alpha}}$ is the sparse coefficient computed by OMP, and $\check{\boldsymbol{\alpha}}$ is the coefficient obtained by ProCRC.

Let $\mathbf{L}=[\boldsymbol{l}_1,\boldsymbol{l}_2,\ldots,\boldsymbol{l}_n]\in\mathbb{R}^{C\times n}$ be the label matrix of the training data, and $\boldsymbol{l}_j=[0,0,\ldots,1,\ldots,0,0]^T\in\mathbb{R}^{C\times 1}$ denotes the label vector of the $j$-th training sample. For the $i$-th class, $\mathbf{L}$ consists of $n_i$ non-zero elements in its $i$-th row, at the indices associated with the columns of $\mathbf{X}_i$. Remember that $\mathbf{X}_i$ is the subset of dictionary atoms belonging to the $i$-th class. Therefore, the $i$-th entry of the vector $\boldsymbol{q}=\mathbf{L} \stackrel{\circ}{\boldsymbol{\alpha}}$ expresses the sum of coefficients in $\stackrel{\circ}{\boldsymbol{\alpha}}$ which correspond to the atoms in $\mathbf{X}_i$, and $\boldsymbol{q}$ is dubbed as the score of each class. Consequently, the test sample is designated into the class that leads to the largest score.

Our proposed SA-ProCRC has the following procedures. Firstly, the dense coefficient and sparse coefficient are obtained by solving (\ref{eq:obj_procrc}) and (\ref{eq:obj_omp}), respectively. Secondly, the dense coefficient is augmented by the sparse coefficient. Finally, the test sample is recognized according to the augmented coefficient vector and the label matrix of the training data. Algorithm \ref{alg1} presents our proposed scheme.

\begin{algorithm}[t]
\begin{algorithmic}
\vspace{0.03in}
\STATE \textbf{Input:} Training data matrix $\mathbf{X}=\left[\mathbf{X}_{1}, \mathbf{X}_{2}, \cdots, \mathbf{X}_{C}\right] \in \mathbb{R}^{m \times n}$ and label matrix $\boldsymbol{L}$, test data $\boldsymbol{y} \in \mathbb{R}^{m}$, parameters $\lambda$ and $\gamma$ for ProCRC, sparsity level $k$ for SRC.
\STATE \textbf{Output:} $\textrm{label}(\boldsymbol{y})=\arg \max _{i}\left(\boldsymbol{q}_{i}\right)$
\STATE \ \ \ \ 1. Compute the coefficient $\check{\boldsymbol{\alpha}}$ of ProCRC by using (\ref{eq:solu_procrc})
\STATE \ \ \ \ 2. Obtain the sparse coefficient $\hat{\boldsymbol{\alpha}}$ of SRC by solving (\ref{eq:obj_omp})
\STATE \ \ \ \ 3. Compute the augmented coefficient $\stackrel{\circ}{\boldsymbol{\alpha}}=\frac{\hat{\boldsymbol{\alpha}}+\check{\boldsymbol{\alpha}}}{\left\|\hat{\boldsymbol{\alpha}}+\check{\boldsymbol{\alpha}}\right\|_{2}}$
\STATE \ \ \ \ 4. Compute $\boldsymbol{q}=\mathbf{L} \stackrel{\circ}{\boldsymbol{\alpha}}$
\vspace{0.03in}
\end{algorithmic}
\caption{SA-ProCRC}
\label{alg1}
\end{algorithm}

\section{Analysis of SA-ProCRC}
\begin{figure}
\centering
\includegraphics[trim={0mm 0mm 0mm 0mm},clip,width=.48\textwidth]{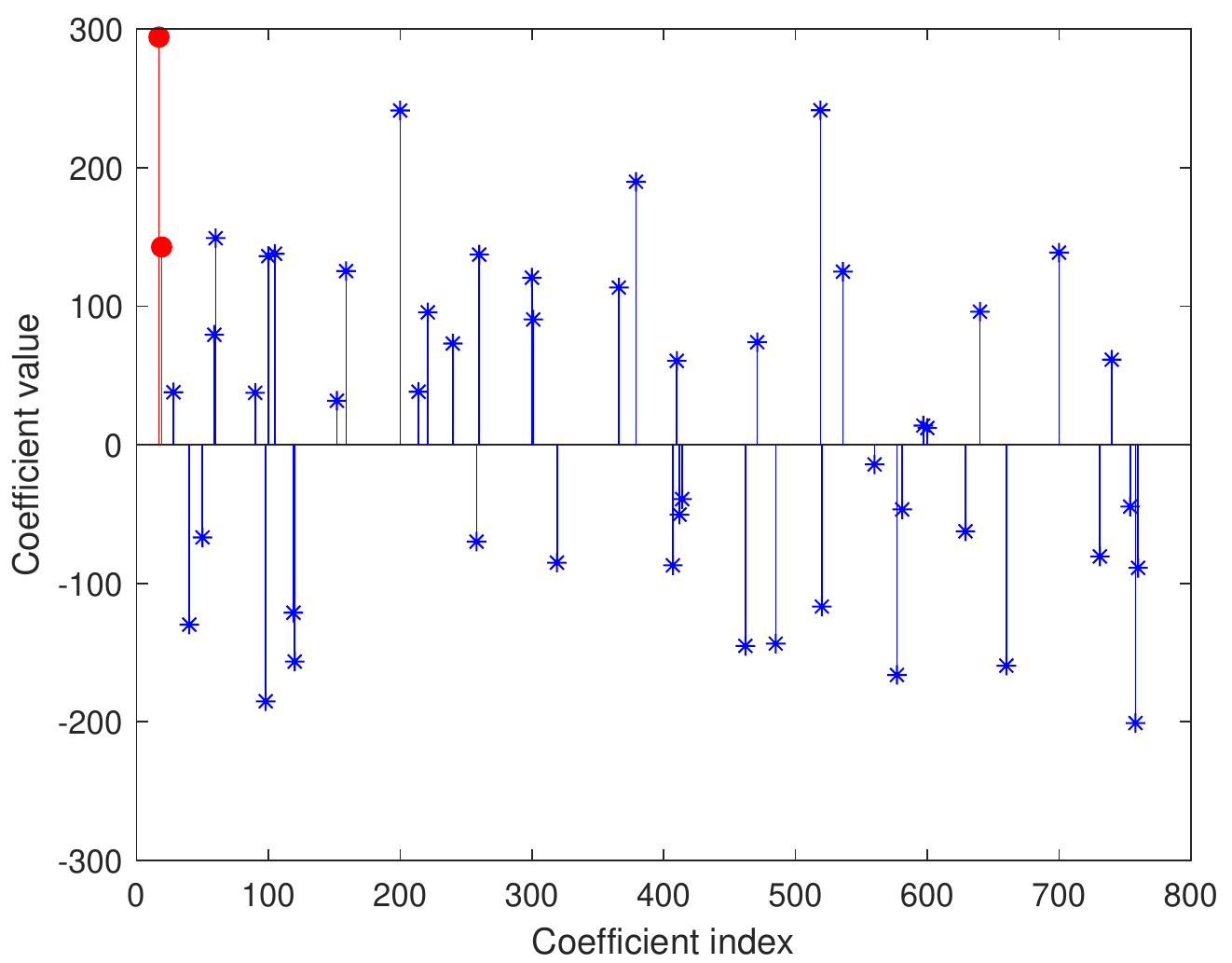}
\caption{Coefficients obtained by SRC.\label{fig:coeff_src}}
\end{figure}

\begin{figure}
\centering
\includegraphics[trim={0mm 0mm 0mm 0mm},clip,width=.48\textwidth]{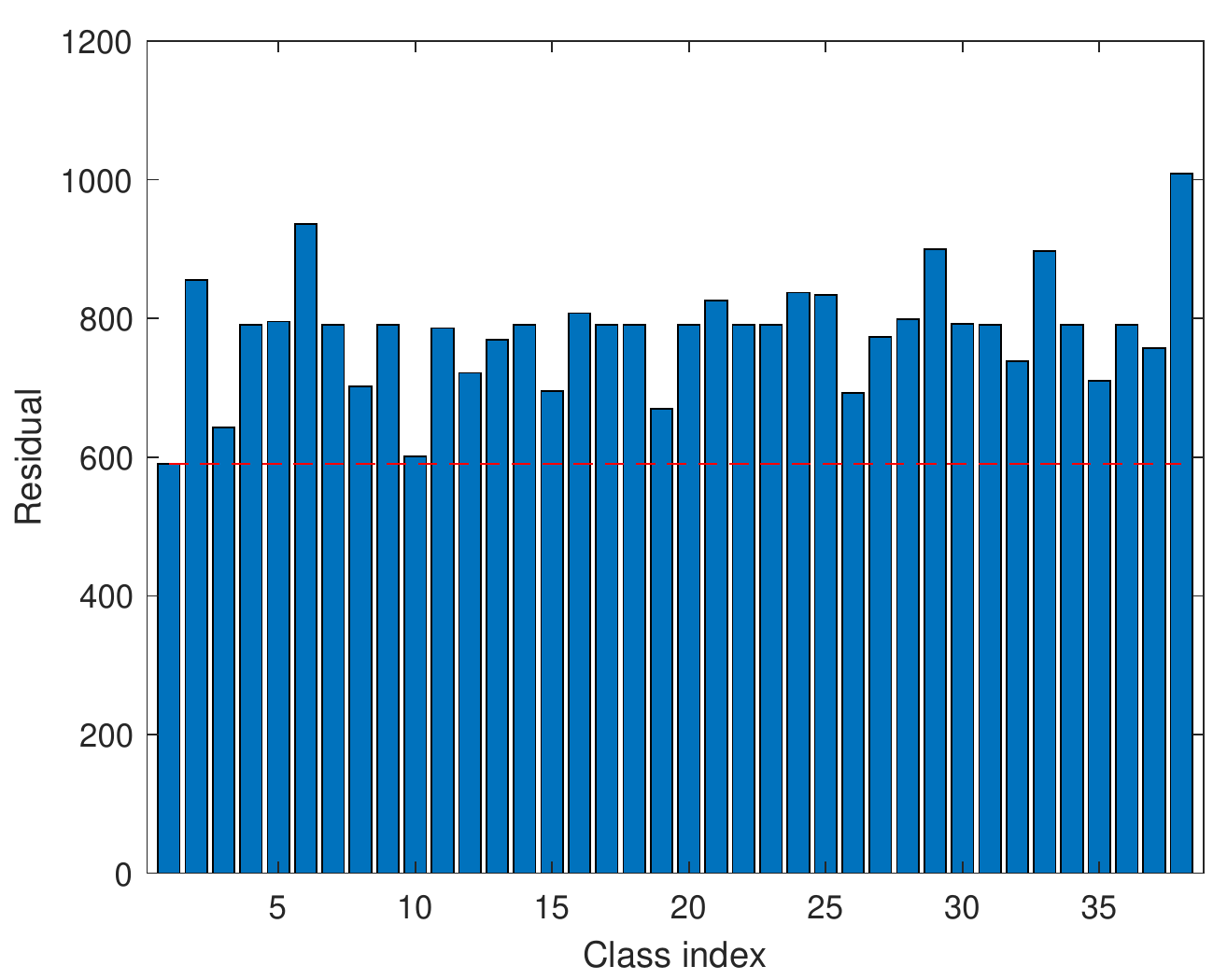}
\caption{The residual of SRC for each class, and the first class has the least residual.\label{fig:res_src}}
\end{figure}

\begin{figure}
\centering
\includegraphics[trim={0mm 0mm 0mm 0mm},clip,width=.48\textwidth]{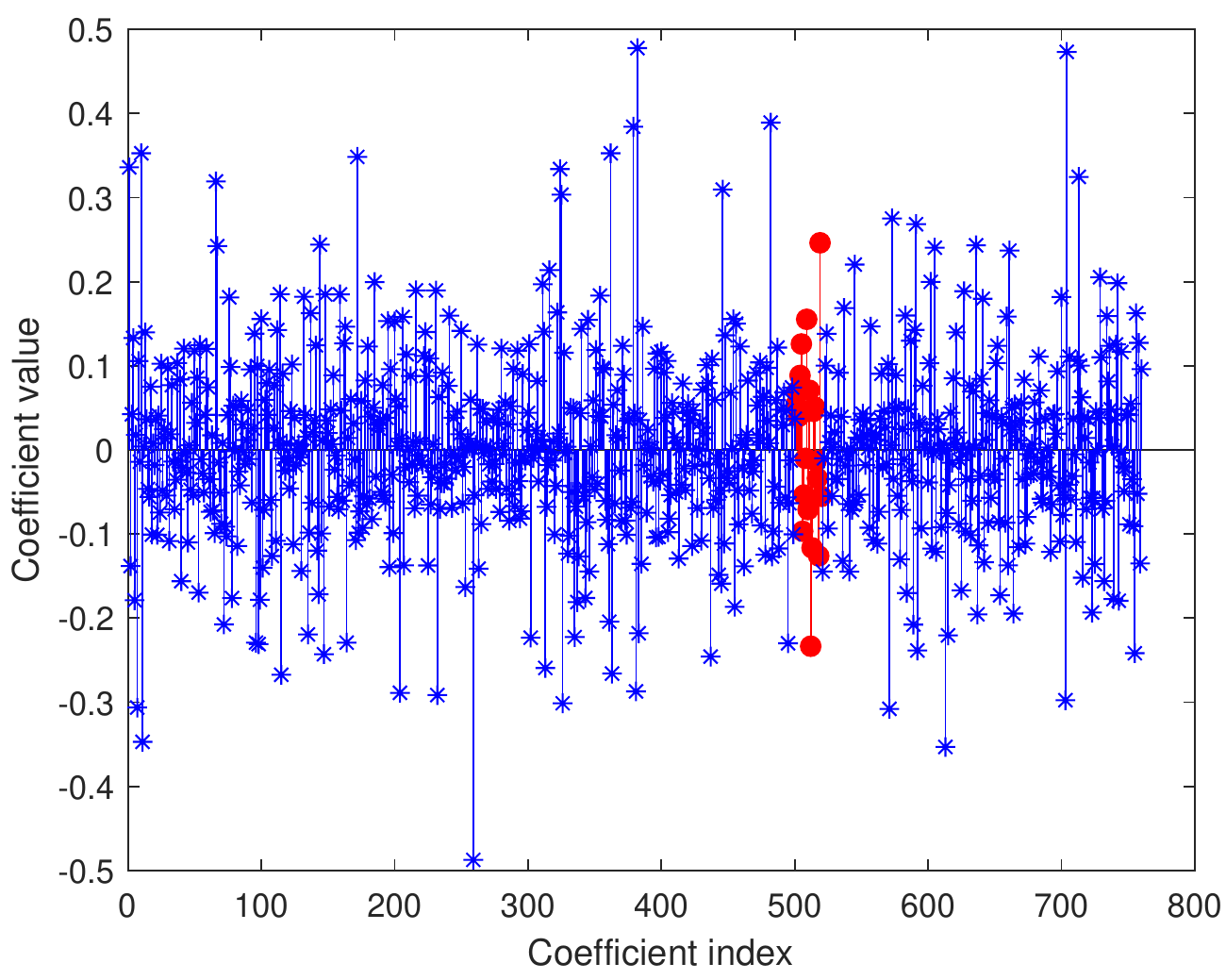}
\caption{Coefficients computed by ProCRC.\label{fig:coeff_procrc}}
\end{figure}

\begin{figure}
\centering
\includegraphics[trim={0mm 0mm 0mm 0mm},clip,width=.48\textwidth]{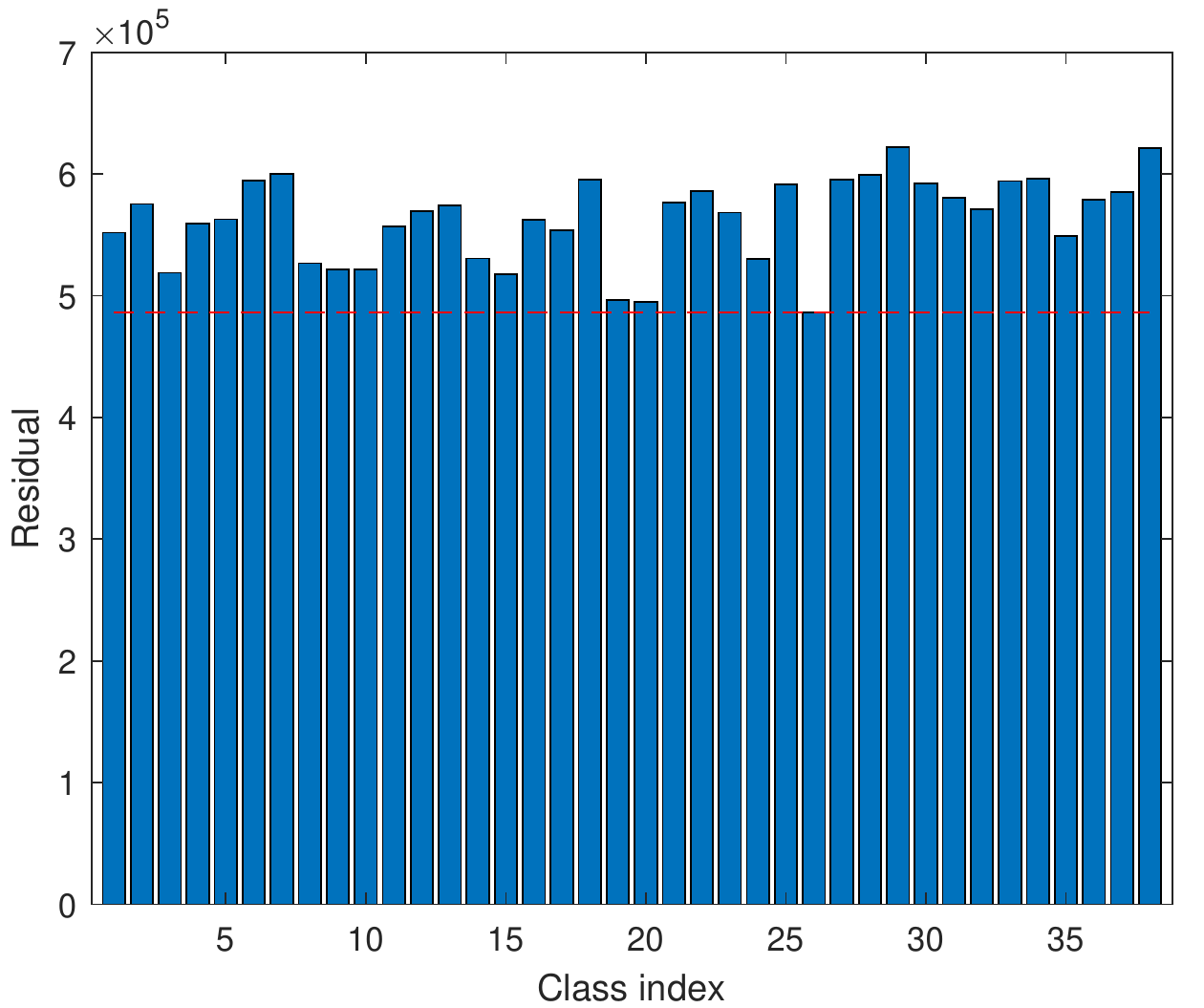}
\caption{The residual of ProCRC for each class, one can see that the 26th class has the minimal residual.\label{fig:res_procrc}}
\end{figure}

\begin{figure}
\centering
\includegraphics[trim={0mm 0mm 0mm 0mm},clip,width=.48\textwidth]{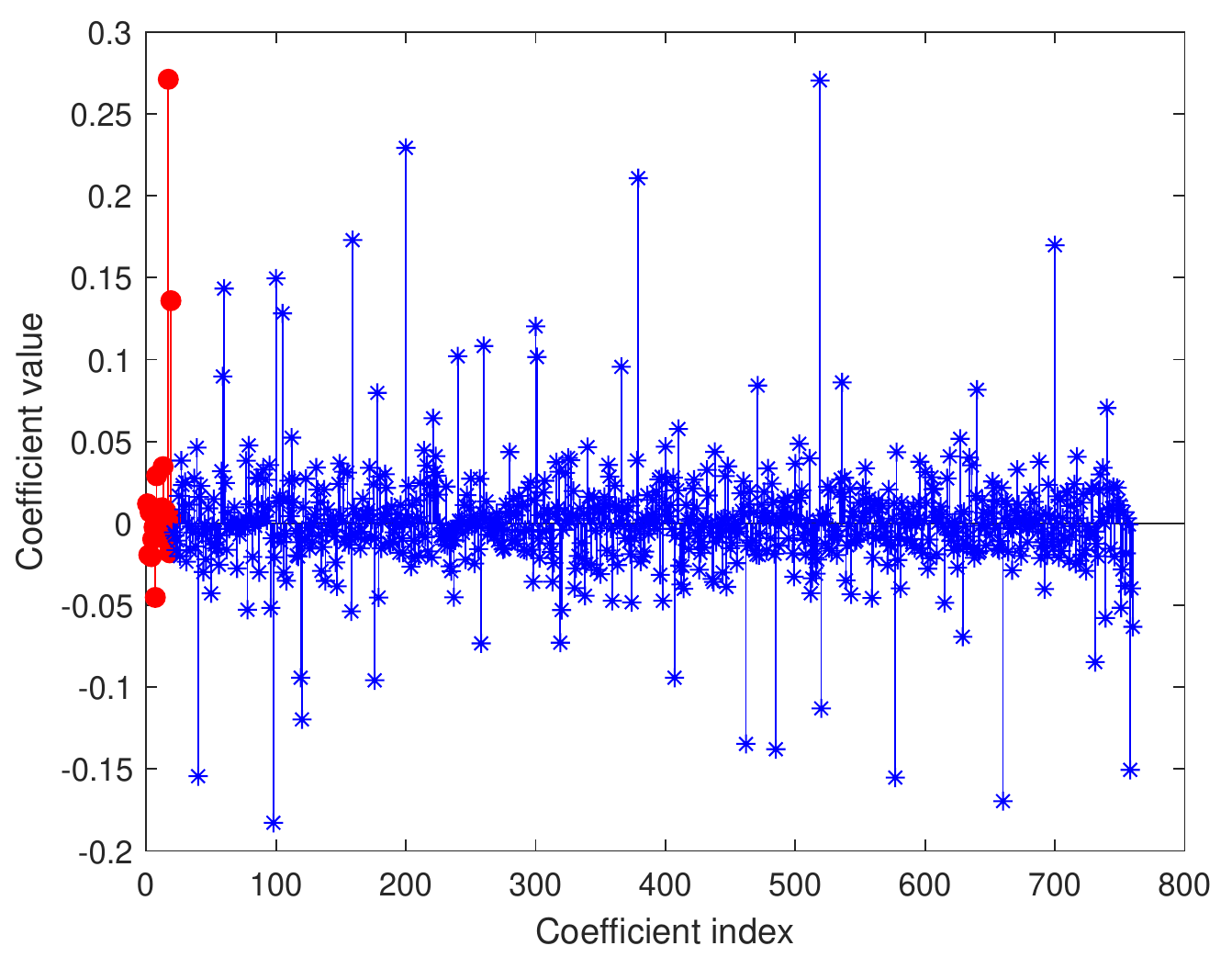}
\caption{Coefficients obtained by SA-ProCRC.\label{fig:coeff_pro}}
\end{figure}

\begin{figure}
\centering
\includegraphics[trim={0mm 0mm 0mm 0mm},clip,width=.48\textwidth]{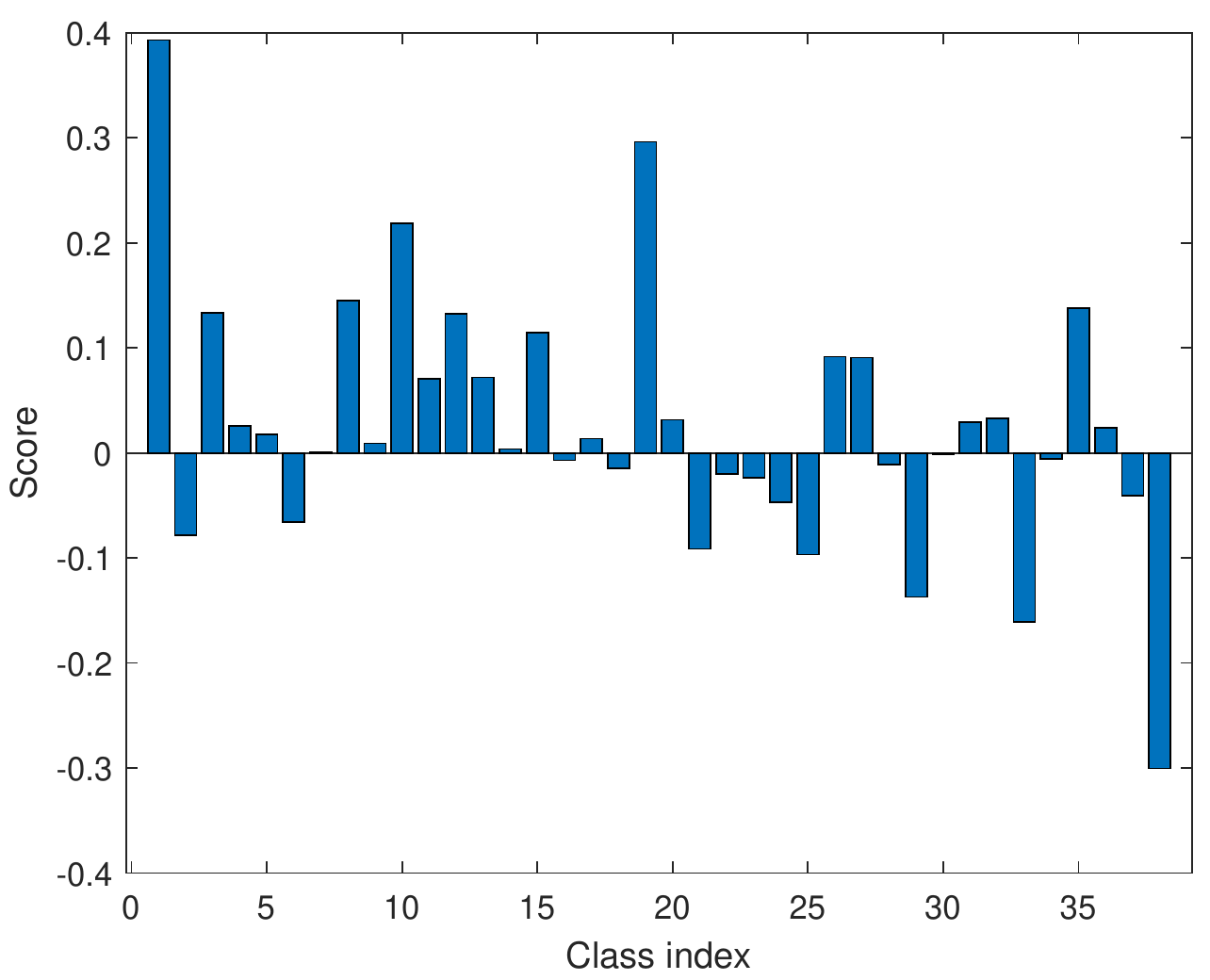}
\caption{The score of SA-ProCRC for each class, it is evident that the first class has the largest value.\label{fig:score_pro}}
\end{figure}

In this section, we present some experimental results on the Extended Yale B database to illustrate the effectiveness of SA-ProCRC. The Extended Yale B database contains 38 individuals and there are about 64 images for each individual. We randomly select 20 images per subject as the training data; therefore, the dictionary contains 760 atoms. We select a test image which belongs to the first subject, and the sparse coefficients and corresponding residual for each class are plotted in Figs.~\ref{fig:coeff_src} and~\ref{fig:res_src}. It can be seen from Fig.~\ref{fig:coeff_src} that coefficients belong to the first class are prominent. From Fig.~\ref{fig:res_src}, we can clearly see that the first class has the least residual, which indicates that the test sample is correctly classified by SRC. Fig.~\ref{fig:coeff_procrc} shows the coefficients derived by ProCRC, we can see that the coefficients are rather dense. Fig.~\ref{fig:res_procrc} presents the residual of ProCRC, one can see that the 26th class has the least residual, thus the test sample is wrongly classified to the 26th class. Coefficients obtained by SA-ProCRC are shown in Fig.~\ref{fig:coeff_pro}, we can see that coefficients from the first class are dominant. Fig.~\ref{fig:score_pro} plots the score of SA-ProCRC for each class, it can be seen that the first class delivers the largest value. As a result, the test sample is designated to the first class by SA-ProCRC. From the above experimental results, we can find that the dense representation of ProCRC may lead to misclassification. By augmenting the dense representation with a sparse representation, the misclassification can be alleviated. This validates the superiority of our proposed SA-ProCRC.

\section{Experiments}
In this section, we conduct experiments on four benchmark datasets: the Yale database, the Extended Yale B database, the AR database and the Scene 15 dataset, {\color{red}the details of these datasets are listed in Table~\ref{table:data_desp}.} We compare the proposed method with state-of-the-art representation based classification methods and several dictionary learning approaches, such as SRC~\cite{R1}, CRC~\cite{R14}, ProCRC~\cite{R15}, D-KSVD~\cite{R17}, LC-KSVD~\cite{R18}, FDDL~\cite{R19}, COPAR~\cite{R20}, JBDC~\cite{R21} and SA-CRC~\cite{R13}. For SRC, we solve the problem in Eq. (\ref{eq:obj_src}) as in Ref.~\cite{R1}. For CRC, LC-KSVD, FDDL, COPAR, JBDC and SA-CRC, we use the publicly available codes. We adapted the code of LC-KSVD to implement D-KSVD. For SA-CRC and our proposed SA-ProCRC, OMP is utilized to obtain the sparse representation. We utilize the same value of sparsity level ($k$=50) as in SA-CRC~\cite{R13}. All experiments are run with MATLAB R2019a under Windows 10 on PC equipped with 3.60 GHz CPU and 16 GB RAM.

\begin{table}[]
\caption{{\color{red}Details of datasets used in our experiments. The columns from left to right are the names of datasets, total number of samples, number of classes and the dimensionality of features.}}
\label{table:data_desp}
\centering
\begin{tabular}{cccc}
\hline
Dataset    & \# Sample & \# Class  & \# Feature \\ \hline
Yale       & 165       & 15     & 576        \\
EYaleB & 2414      & 38     & 504        \\
AR         & 2600      & 100    & 540        \\
Scene 15   & 4485      & 15     & 3000       \\ \hline
\end{tabular}
\end{table}

\subsection{Experiments on the Yale database}
\begin{figure}
\centering
\includegraphics[trim={0mm 0mm 0mm 0mm},clip,width=.48\textwidth]{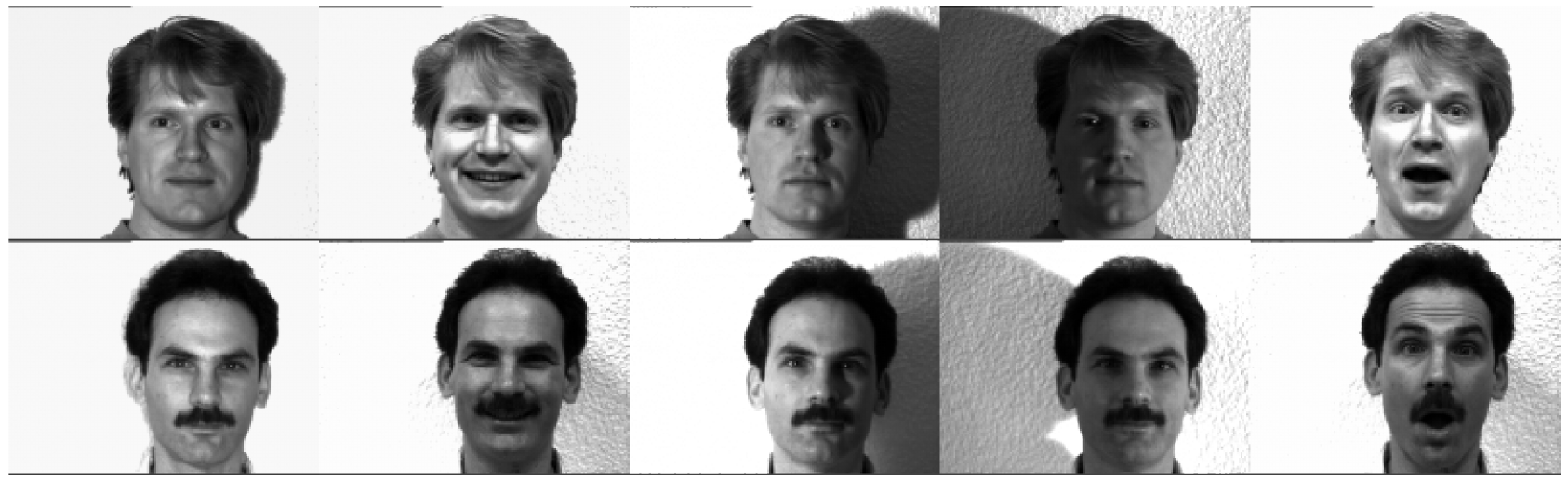}
\caption{Example images from the Yale database.\label{fig:exam_Yale}}
\end{figure}

There are 165 images for 15 subjects in the Yale database, each has 11 images. These images have illumination and expression variations, Fig.~\ref{fig:exam_Yale} shows some example images from this database. All the images are resized to 24$\times$24 pixels, leading to a 576-dimensional vector. In our experiments, six images per subject are randomly selected for training and the rest for testing. The error tolerance $\varepsilon$ of SRC is 0.05, and the balancing parameter $\lambda$ of CRC is 0.001. The sparsity level and number of atoms for D-KSVD and LC-KSVD are 30 and 60, respectively. Sparsity level $k$ and $\lambda$ of SA-CRC are set to be 50 and 0.002, respectively.
Experimental results are summarized in Table~\ref{table:acc_yale}, in which the best result is highlighted by bold number. It can be observed that SA-ProCRC achieves the highest recognition accuracy, with a 17\% reduction in the error rate of ProCRC, and 12\% reduction in that of SA-CRC.

\begin{table}[h]
\small\sf\centering
\caption{Recognition accuracy on the Yale database.\label{table:acc_yale}}
\begin{tabular}{lll}
\toprule
Methods & Accuracy (\%)  \\
\midrule
SRC& 95.06$\pm$3.32\\
CRC &94.53$\pm$2.97\\
ProCRC &95.33$\pm$2.82 \\
D-KSVD & 94.26$\pm$2.88\\
LC-KSVD & 94.53$\pm$0.03\\
FDDL & 95.73$\pm$3.00\\
COPAR &91.33$\pm$4.23 \\
JBDC & 94.93$\pm$2.72\\
SA-CRC & 95.60$\pm$2.59\\
SA-ProCRC &\textbf{96.13$\pm$2.84} \\
\bottomrule
\end{tabular}\\[10pt]
\end{table}

\subsection{Experiments on the Extended Yale B database}
\begin{figure}
\centering
\includegraphics[trim={0mm 0mm 0mm 0mm},clip,width=.48\textwidth]{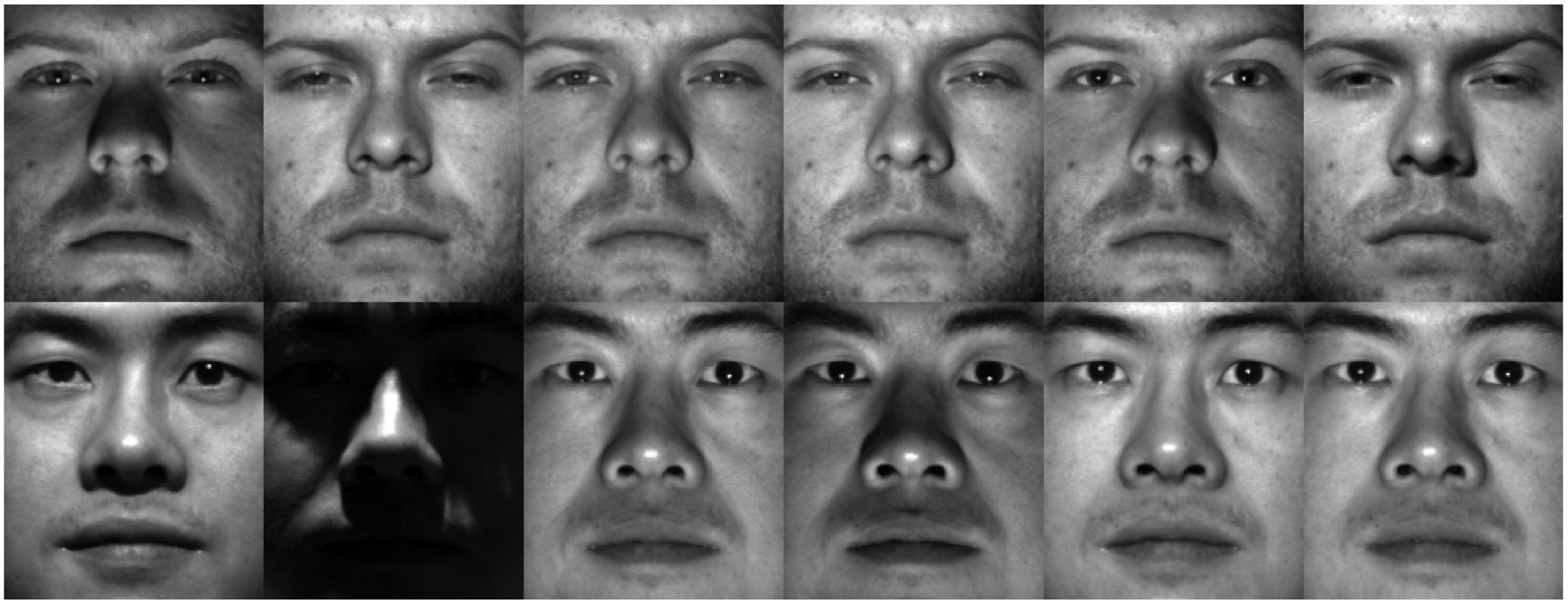}
\caption{Example images from the Extended Yale B database.\label{fig:exam_EYaleB}}
\end{figure}

The Extended Yale B face database is composed of 2414 images of 38 individuals. Each individual has 59-64 images taken under different illumination conditions, example images from this dataset are shown in Fig.~\ref{fig:exam_EYaleB}. In our experiments, each 192$\times$168 image is projected onto a 504-dimensional space via random projection. 20 images per person are selected for training and the remaining for testing. We use the error tolerance of 0.05 for SRC, and the regularization parameter $\lambda$=0.001 for CRC. The sparsity level and number of atoms for D-KSVD and LC-KSVD are 50 and 400, respectively. Sparsity level $k$ and $\lambda$ of SA-CRC are set to be 50 and 0.005, respectively. Table~\ref{table:acc_yaleb} lists the recognition accuracy of the comparison methods. It can be seen that our proposed SA-ProCRC is superior to its competing approaches.

\begin{table}[h]
\small\sf\centering
\caption{Recognition accuracy on the Extended Yale B database.\label{table:acc_yaleb}}
\begin{tabular}{lll}
\toprule
Methods & Accuracy (\%)  \\
\midrule
SRC& 93.18$\pm$0.55\\
CRC &94.77$\pm$0.48\\
ProCRC &94.82$\pm$0.49 \\
D-KSVD & 90.79$\pm$0.51\\
LC-KSVD & 91.48$\pm$0.69\\
FDDL & 92.32$\pm$0.68\\
COPAR &90.81$\pm$0.55 \\
JBDC & 94.74$\pm$0.83\\
SA-CRC & 95.52$\pm$0.73\\
SA-ProCRC &\textbf{95.64$\pm$0.78} \\
\bottomrule
\end{tabular}\\[10pt]
\end{table}

\subsection{Experiments on the AR database}
\begin{figure}
\centering
\includegraphics[trim={0mm 0mm 0mm 0mm},clip,width=.48\textwidth]{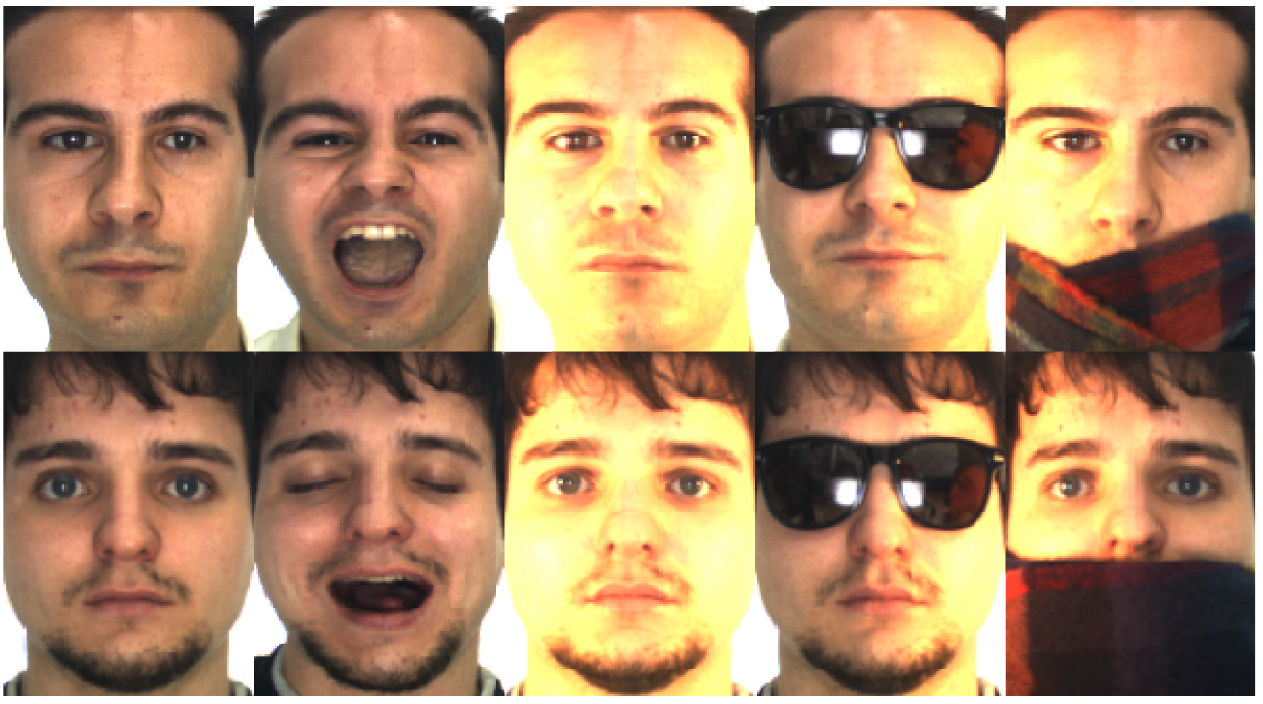}
\caption{Example images from the AR database.\label{fig:exam_AR}}
\end{figure}

The AR database has more than 4000 face images of 126 subjects with variations in facial expression, illumination conditions and occlusions, Fig.~\ref{fig:exam_AR} shows example images from this database. We use a subset of 2600 images of 50 male and 50 female subjects from the database. Each 165$\times$120 face image is projected onto a 540-dimensional vector by random projection. 10 images per person are randomly selected for training and the remaining for testing. The error tolerance of SRC is 0.05, and the balancing parameter of CRC is 0.0014. The sparsity level and number of atoms for D-KSVD and LC-KSVD are 50 and 600, respectively. Sparsity level $k$ and $\lambda$ of SA-CRC are set to be 50 and 0.002, respectively. Experimental results are shown in Table~\ref{table:acc_ar}. We can see that the best classification result is achieved by our proposed SA-ProCRC, with a 23\% reduction in the error rate of ProCRC.

\begin{table}[h]
\small\sf\centering
\caption{Recognition accuracy on the AR database.\label{table:acc_ar}}
\begin{tabular}{lll}
\toprule
Methods & Accuracy (\%)  \\
\midrule
SRC& 91.25$\pm$1.17\\
CRC &92.04$\pm$0.83\\
ProCRC &93.03$\pm$0.64 \\
D-KSVD & 90.31$\pm$1.13\\
LC-KSVD & 89.31$\pm$1.27\\
FDDL & 91.01$\pm$0.99\\
COPAR &89.06$\pm$1.54 \\
JBDC & 90.97$\pm$0.79\\
SA-CRC & 93.74$\pm$0.84\\
SA-ProCRC &\textbf{94.67$\pm$0.66} \\
\bottomrule
\end{tabular}\\[10pt]
\end{table}

\subsection{Experiments on the Scene 15 dataset}
\begin{figure}
\centering
\includegraphics[trim={0mm 0mm 0mm 0mm},clip,width=.48\textwidth]{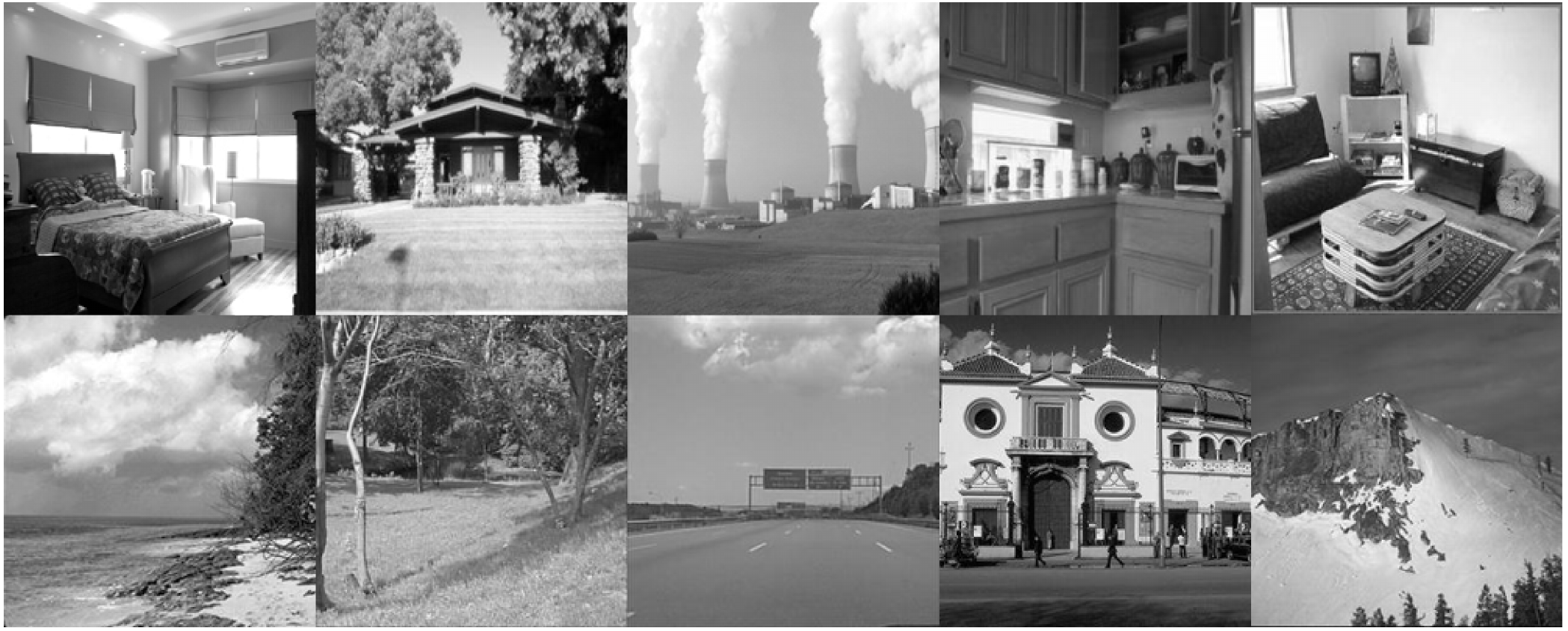}
\caption{Example images from the Scene 15 dataset.\label{fig:exam_s15}}
\end{figure}

This dataset contains 15 natural scene categories including a wide range of indoor and outdoor scenes, such as bedroom, office and mountain, example images from this dataset are shown in Fig.~\ref{fig:exam_s15}. For fair comparison, we employ the 3000-dimensional SIFT-based features used in LC-KSVD ~\cite{R18}. We randomly select 50 images per category as training data and use the rest for testing. The error tolerance of SRC is 1e-6, and the balancing parameter of CRC is 1. 50 atoms are used for D-KSVD and LC-KSVD. Sparsity level $k$ and $\lambda$ of SA-CRC are set to be 50 and 1, respectively. Recognition accuracy of different approaches on this dataset are presented in Table~\ref{table:acc_scene15}. Again, SA-ProCRC outperforms the comparison methods.

\begin{table}[h]
\small\sf\centering
\caption{Recognition accuracy on the Scene 15 dataset.\label{table:acc_scene15}}
\begin{tabular}{lll}
\toprule
Methods & Accuracy (\%)  \\
\midrule
SRC& 95.41$\pm$0.13\\
CRC &96.15$\pm$0.33\\
ProCRC &96.56$\pm$0.35 \\
D-KSVD & 95.12$\pm$0.18\\
LC-KSVD & 96.37$\pm$0.28\\
FDDL & 94.08$\pm$0.43\\
COPAR &96.02$\pm$0.28 \\
JBDC & 97.36$\pm$0.32\\
SA-CRC & 97.18$\pm$0.25\\
SA-ProCRC &\textbf{97.56$\pm$0.20} \\
\bottomrule
\end{tabular}\\[10pt]
\end{table}

\section{Conclusions}
It has been argued that it is the collaborative representation mechanism rather that the sparsity constraint that makes SRC powerful for pattern classification. As a result, sparsity is ignored to some extent in CRC and its extensions. To address this problem, we present a sparsity augmented probabilistic collaborative representation based classification (SA-ProCRC) method to promote the sparsity in ProCRC. The proposed SA-ProCRC is computationally efficient due to the fact that ProCRC has closed-form solution. Meanwhile, discriminative information contains in the resulting sparse coefficient can be exploited in SA-ProCRC. In essence, SA-ProCRC is a classifier, thus it can be applied to other pattern classification tasks. In our future work, we will evaluate SA-ProCRC with deep features and develop new representation based classification algorithm.

\begin{acks}
The authors would like to thank Prof. Naveed Akhtar for providing the source code of SA-CRC at \url{http://staffhome.ecm.uwa.edu.au/~00053650/code.html}.
\end{acks}

\begin{dci}
The author(s) declared no potential conflicts of interest with respect to the research, authorship, and/or publication of this article.
\end{dci}

\begin{funding}
The author(s) disclosed receipt of the following financial support for the research, authorship, and/or publication of this article: This work was supported by the National Natural Science Foundation of China (Grant No. 61672265).
\end{funding}

\end{document}